\documentclass{article}
\usepackage{ijcai24}

\usepackage{times}
\usepackage{soul}
\usepackage{url}
\usepackage[hidelinks]{hyperref}
\usepackage[utf8]{inputenc}
\usepackage[small]{caption}
\usepackage{graphicx}
\usepackage{amsmath}
\usepackage{amsthm}
\usepackage{booktabs}
\usepackage{algorithm}
\usepackage[switch]{lineno}
\usepackage{xspace}
\usepackage{bm} 
\usepackage{multirow}
\usepackage{array}
\usepackage{booktabs}
\usepackage{amsfonts}
\usepackage{amsmath}
\usepackage{amsthm}
\usepackage{subfigure}
\usepackage{listings}
\usepackage{threeparttable}
\usepackage{algorithm}
\usepackage{algpseudocode}
\usepackage{diagbox}
\usepackage{pifont}
\usepackage{graphicx}
\usepackage{amssymb}
\usepackage{tcolorbox}

\definecolor{pink}{rgb}{0.858, 0.188, 0.478}
\definecolor{green}{rgb}{0.2, 0.5, 0.15}
\definecolor{orange}{rgb}{0.97, 0.6, 0.2}
\definecolor{purple}{rgb}{0.6, 0.4, 0.7}
\definecolor{darkoragne}{rgb}{1.0, 0.549, 0.0}
\definecolor{lightoragne}{rgb}{1.0, 0.5, 0.9}
\definecolor{darkred}{rgb}{0.545, 0, 0}
\definecolor{skyblue}{rgb}{0.000, 0.000, 0.902}
\definecolor{blue2}{rgb}{0.047, 0.365, 0.647}
\definecolor{red2}{rgb}{0.863, 0.075, 0.235}
\definecolor{gray}{rgb}{0.620, 0.620, 0.620}
\definecolor{pink}{rgb}{1.0, 0.4, 0.4}

\definecolor{c1}{rgb}{1.0, 0.9, 0.9}
\definecolor{c2}{rgb}{1.0, 0.7, 0.7}
\definecolor{c3}{rgb}{1.0, 0.6, 0.6}
\definecolor{c4}{rgb}{1.0, 0.4, 0.4}
\definecolor{c5}{rgb}{1.0, 0.1, 0.1}

\newcommand{\Frst}[1]{\textcolor{pink}{\textbf{#1}}}
\newcommand{\Scnd}[1]{\textcolor{skyblue}{\underline{#1}}}
\definecolor{commentcolor}{RGB}{110,154,155}   % define comment color
\newcommand{\PyComment}[1]{\ttfamily\textcolor{commentcolor}{\# #1}}  % add a "#" before the input text "#1"
\newcommand{\PyCode}[1]{\ttfamily\textcolor{black}{#1}} % \ttfamily is the code font

\newcommand{\relu}{\text{ReLU}\xspace}
\newcommand{\nosection}[1]{\noindent\textbf{#1\ }}
\newcommand{\ours}[0]{BlockGCL\xspace}

\title{Oversmoothing: A Nightmare for Graph Contrastive Learning?}

\author{
Jintang Li$^1$
\and
Wangbin Sun$^1$\and
Ruofan Wu$^2$\and
Yuzhang Zhu$^1$\and
Zibin Zheng$^1$\And
Liang Chen$^1$\\
\affiliations
$^1$Sun Yat-sen University\\
$^2$Ant Group\\
\emails
lijt55@mail2.sysu.edu.cn
}

\begin{document}

\maketitle

\begin{abstract}
    Oversmoothing is a common phenomenon observed in graph neural networks (GNNs), in which an increase in the network depth leads to a deterioration in their performance. Graph contrastive learning (GCL) is emerging as a promising way of leveraging vast unlabeled graph data. As a marriage between GNNs and contrastive learning, it remains unclear whether GCL inherits the same oversmoothing defect from GNNs. This work undertakes a fundamental analysis of GCL from the perspective of oversmoothing on the first hand. We demonstrate empirically that increasing network depth in GCL also leads to oversmoothing in their deep representations, and surprisingly, the shallow ones. We refer to this phenomenon in GCL as \textit{`long-range starvation'}, wherein lower layers in deep networks suffer from degradation due to the lack of sufficient guidance from supervision. Based on our findings, we present \ours, a remarkably simple yet effective blockwise training framework to prevent GCL from notorious oversmoothing. Without bells and whistles, \ours consistently improves robustness and stability for well-established GCL methods with increasing numbers of layers on several real-world graph benchmarks.
\end{abstract}

\section{Introduction}

Graph neural networks (GNNs), following a standard neighborhood aggregation (a.k.a. \textit{message passing}) paradigm, have grown into powerful tools to model non-Euclidean graph-structured data arising from various fields. In recent years, GNNs have garnered increasing interest from the machine learning community and have been widely adopted in many applications, ranging from social network analysis~\cite{LiuCHPZT22} and recommendation systems~\cite{lightgcn} to molecular prediction~\cite{DBLP:conf/ijcai/ZhaoLHLZ21} and drug discovery~\cite{DBLP:journals/corr/abs-1905-00534}.

However, GNNs are known to suffer two fundamental problems which restrict their learning ability on general graph-structured data. Firstly, most of the existing GNNs fail to be `deep enough' due to the oversmoothing issue, where deeply stacking layers give rise to indistinguishable representations and significant performance deterioration~\cite{LiHW18,contranorm}.
Secondly, GNNs require task-dependent annotations or labels to learn meaningful representations, which are extremely scarce, expensive, or even unavailable to obtain in practice~\cite{DBLP:journals/corr/abs-2103-00111,maskgae}.
These obstacles greatly limit the applications of GNNs in real-world scenarios, which in turn have motivated two important lines of research: deep GNNs and graph self-supervised learning.

Deep GNN architectures offer several key advantages, such as larger receptive fields and the ability to capture long-range dependencies, both of which are beneficial in many real-world graphs like molecules~\cite{DBLP:conf/ijcai/ZhaoLHLZ21} and brain networks~\cite{braingb}. As a result, efforts have been dedicated to deepening GNNs, with the goal of achieving better performance and avoiding the issue of oversmoothing.  Typical approaches can be divided into four categories, including regularisation~\cite{dropedge,dropnode}, normalization~\cite{contranorm,nodenorm,pairnorm}, architectural exploration~\cite{appnp,dagnn,deepgcns}, and residual connection~\cite{jknet,gcn2} (see Section~\ref{sec:related_work}). Yet, a significant amount of research on addressing the issue of oversmoothing in GNNs is still limited to supervised or semi-supervised scenarios.

Research into graph self-supervised learning has grown in parallel with the development of deep GNNs.
Graph contrastive learning (GCL) emerges as a promising paradigm for learning graph representations in a self-supervised manner, without requiring human annotations~\cite{maskgae,dgi,bgrl}.
The basic idea is to train an encoder network (typically a GNN) to distinguish between pairs of graph views such that similar (`positive') examples have similar representations and dissimilar (`negative') examples have dissimilar representations.
By leveraging the great ability of GNNs in modeling graph-structured data, GCL has achieved state-of-the-art results in various graph-based learning tasks~\cite{wu2021self}.
As these GCL models become increasingly powerful, it still remains unclear to what extent they are capable of learning meaningful and distinguishable representations by means of deeper architectures. In GCL, we use the term `depth' to refer to the number of layers in the \textit{encoder} network, in order to align with the existing literature on oversmoothing in GNNs.

\textbf{Will GCL present the same oversmoothing as GNN?}
We provide what is to the best of our knowledge the first study regarding the oversmoothing in current GCL methods (surprisingly overlooked).
Our empirical studies demonstrate that GCL, which relies on the learning ability of GNNs, is subject to the same problem of oversmoothing.
We also observe an interesting phenomenon that deep architectures can also lead to oversmoothing in shallow representations - representations learned from lower layers.
We attribute the problem to \textit{`long-range starvation'}.
Although the lower layers play a key role in GCL, training losses are computed at the top layer, and weight updates are computed based on the gradient that flows from the very top.
The lower layers are not explicitly guided and the interaction among them is only used for calculating new activations. This limits the potential of GCLs and prevents them from being scaled to deeper architectures.

\textbf{Can we make vanilla GCL methods go deeper without significantly sacrificing their performance?}
Motivated by the above findings, in this work we seek to address the long-range starvation problem in GCL by a new local training paradigm.
In particular, we propose a simple yet effective blockwise learning paradigm for GCL, which merges one or more consecutive layers of GNNs into several individual blocks, in which a self-defined contrastive loss is applied locally to explicitly guide the learning. In addition, we limit the backpropagation path to a single block during training to avoid excessive memory footprints and minimize the negative effects of individual losses from near blocks. Such a simple design drastically improved the performance of GCLs against oversmoothing.

Our main contributions can be summarized as:
\begin{itemize}
    \item \textbf{Deep GCL models suffer from serious oversmoothing problems inherited from GNNs.}
          We rigorously investigate the oversmoothing phenomenon in well-established GCL models to offer new insights into addressing oversmoothing problems. To our best knowledge, our work is the first to study the oversmoothing problem in GCL.
    \item \textbf{Shallow representations in deep GCL models may collapse without proper guidance, leading to oversmoothing.}
          We have empirically observed an interesting `long-range starvation' phenomenon, in which shallow representations in deep GCL models are collapsed. Surprisingly, the representations of the last few layers, where the contrastive loss is applied, are relatively less affected by oversmoothing.
    \item \textbf{Blockwise learning paradigm helps alleviate the oversmoothing.}
          We introduce a local learning scheme for GCL, \ours, which merges consecutive layers of the network into a single block and independently guides each block for training. In our experiments, \ours makes GCL models significantly more robust to oversmoothing and as a result, enables the training of deeper networks with less comprised in performance.
    \item \textbf{\ours comes with good generality and flexibility for well-established GCL models.}
          Our proposed \ours is a derivative of the standard GCL framework with minimal but effective modifications on the training paradigm, which is very straightforward to implement and is not specific to any particular GCL models but rather applies broadly.
\end{itemize}

\section{Related Work}
\label{sec:related_work}
In this section, we will briefly review the recent advances regarding the GCL research, and then describe the literature on oversmoothing of GNNs that is closely related to this work.

\nosection{Graph contrastive learning.}
GCL is a prevalent graph self-supervised learning paradigm, which encourages an
encoder (typically a GNN) to learn useful representations by maximizing the mutual information (MI) between instances. As a pioneer work, DGI~\cite{dgi} firstly extends Deep InfoMax~\cite{dim} to graphs by maximizing the correspondence of local node representations and summarized graph representations. GRACE~\cite{grace} and MVGRL~\cite{mvgrl} then proceed to learn node representations by contrasting two graph augmentation views, pulling the representation of the same node in two views close.
Despite these achievements, most GCL methods have limitations such as time-consuming training, memory inefficiency, and poor scalability~\cite{ggd}. Recent efforts have been made for scaling up GCL through architecture simplification (e.g., BGRL~\cite{bgrl} and GGD~\cite{ggd}) or in-batch feature decorrelation (e.g., CCA-SSG~\cite{cca_ssg}).
However, whether GCL would suffer from the oversmoothing issue is, as far as we know, still an open question.

\nosection{Oversmoothing in GNNs.}
Oversmoothing is a notorious issue in deep GNNs, and many techniques have been proposed to alleviate it practically. The oversmoothing issue is firstly observed and raised by Li et al.~\cite{LiHW18}. Theoretically, node representations gradually converge to a constant after repeatedly exchanging messages with neighbors when the layer goes to infinity~\cite{OonoS20}.
As an active area of research, current works have explored different approaches to resolve the difficulties in training deep GNNs. Firstly, regularization or normalization techniques such as DropEdge~\cite{dropedge}, DropNode~\cite{dropnode}, DropConnect~\cite{dropconnect}, NodeNorm~\cite{nodenorm}, PairNorm~\cite{pairnorm}, and ContraNorm~\cite{contranorm} are proposed as plug-and-play techniques to prevent features from becoming indistinguishable as propagating.
Another line of research~\cite{appnp,dagnn,deepergcn} has progressed to efficient propagation schemes to solve the oversmoothing problem.
In addition, skip connection~\cite{jknet}, residual connections~\cite{gcn2}, and reversible connections~\cite{deep1000} are widely adopted as empirical effective designs in training deeper GNNs.
Note that \ours is designed to address oversmoothing from the perspective of \textit{training paradigm}, which is orthogonal to the above advances in learning deep GNNs.

\section{Background}
\nosection{Notations.}
We first establish common notations for working with graphs.
Let $\mathcal{G}=(\mathcal{V}, \mathcal{E})$ be an undirected graph with nodes $\mathcal{V}$ and edges $\mathcal{E}$. We denote $\mathbf{X}\in\mathbb{R}^{N\times F}$ the feature matrix with $N=|\mathcal{V}|$ the number of nodes and $F$ the feature dimension. The goal of self-supervised learning is to learn useful node representations $\mathbf{Z}\in\mathbb{R}^{N\times D}$ with self-defined supervision, where $D$ is the hidden dimension.

\nosection{Graph neural networks.}
The vast majority of GNNs follow the canonical \textit{message passing} scheme in which each node's representation is computed recursively by aggregating representations (`messages') from its immediate neighbors~\cite{kipf2016semi,hamilton2017inductive}.
The message passing scheme is tailored by two core functions, i.e., a message aggregation function $\operatorname{AGGREGATE}$, and an update function $\operatorname{UPDATE}$.
Let $h_u^l$ denote the embedding of node $u$ at the $l$-th layer of GNN, where $l=\{1,\dots,L\}$ and initially $h^0_{u}=x_{u}$. The propagation at each layer $l$ has two main steps: (i) \textbf{Message aggregation}. For each node pair $(u, v)$ that is associated with a linked edge, the message received by a node $u$ is aggregated from its direct neighbors using propagated message at the current layer $l$. (ii) \textbf{Embedding update.} When received all messages, a non-linear function $\operatorname{UPDATE}$ is adopted to update node embeddings for each node $u$. In general, the update of node $u$'s representation in $l$-th layer is formalized as follows:
\begin{align}
    h_{\mathcal{N}_u}^{(l)} & = \operatorname{AGGREGATE}^{(l)}\left(\left\{h_{v}^{(l-1)}: v \in \mathcal{N}_u\right\}\right), \\
    h_u^{(l)}               & =\operatorname{UPDATE}^{(l)}\left(\left\{h_u^{(l-1)}, h_{\mathcal{N}_u}^{(l)}\right\}\right),
\end{align}
where $\mathcal{N}(u)$ denotes the neighborhood set of node $u$.
Overall, the final node representation is the output of the $L$-th layer, i.e., $z_u=h_u^L, \forall u \in \mathcal{V}$.

\nosection{Graph contrastive learning.}
A general GCL framework typically leverages a dual-branch architecture with three key components: (i) augmentation, (ii) contrastive networks, and (iii) contrastive objective.
At each iteration of training, two graph views are generated from input graph $\mathcal{G}$ via stochastic augmentation functions $t_A, t_B \sim \mathcal{T}$ such that $\mathcal{G}_A=t_A(\mathcal{G})$ and $\mathcal{G}_B=t_B(\mathcal{G})$, where $\mathcal{T}$ is the set of all possible augmentations (e.g., edge dropping and feature masking~\cite{cca_ssg}). Then, the representations in two branches are obtained by a shared GNNs network $f_\theta$, denoted by $\mathbf{Z}_A=f_\theta(\mathcal{G}_A)$, $\mathbf{Z}_B=f_\theta(\mathcal{G}_B)$ where $ \mathbf{Z}_A, \mathbf{Z}_B \in \mathbb{R}^{N \times D}$. Finally, a contrastive loss is employed to measure the similarity of positive samples and the discrepancy between negatives.

\section{Preliminary Study}

In this section, we aim to study how GCL behaves under deep networks. Specifically, we first conduct empirical study on several GCL methods to evaluate their performance w.r.t. increasing depth of networks, and then and investigate the underlying reasons for the phenomenon.

\subsection{Oversmoothing in GCL}
\label{sec:empirical_1}

GCL works by maximizing the mutual information of different graph views to learn \textbf{discriminative} node representations. This stands in contrast to the phenomenon of oversmoothing, where node representations become increasingly similar and difficult to distinguish from one another. Therefore, it is intuitive that GCL would perform better in situations where oversmoothing occurs.

To investigate how GCL methods behave with deep architectures, we design node classification experiments on two common datasets, i.e., Cora and Citeseer~\cite{sen2008collective}, whose statistics are summarized in Table~\ref{tab:dataset}. The experiments are performed on supervised GCN~\cite{kipf2016semi} and three well-established GCL methods: DGI~\cite{dgi}, GRACE~\cite{grace}, and CCA-SSG~\cite{cca_ssg}.
All methods follow the hyperparameter settings in their original papers.
All of the GCL methods use GCN as the encoder network for a fair comparison.
We report the average performance over 5 runs.

\begin{figure}[t]

    \centering
    \includegraphics[width=0.46\linewidth]{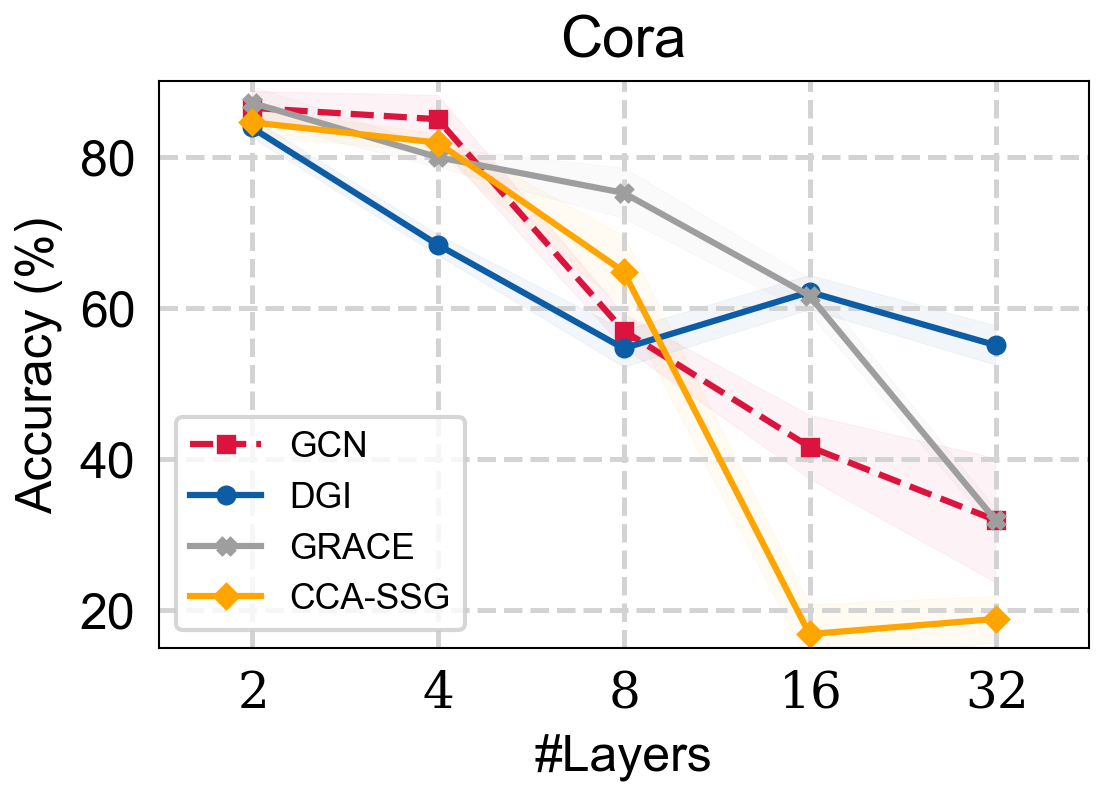}
    \includegraphics[width=0.46\linewidth]{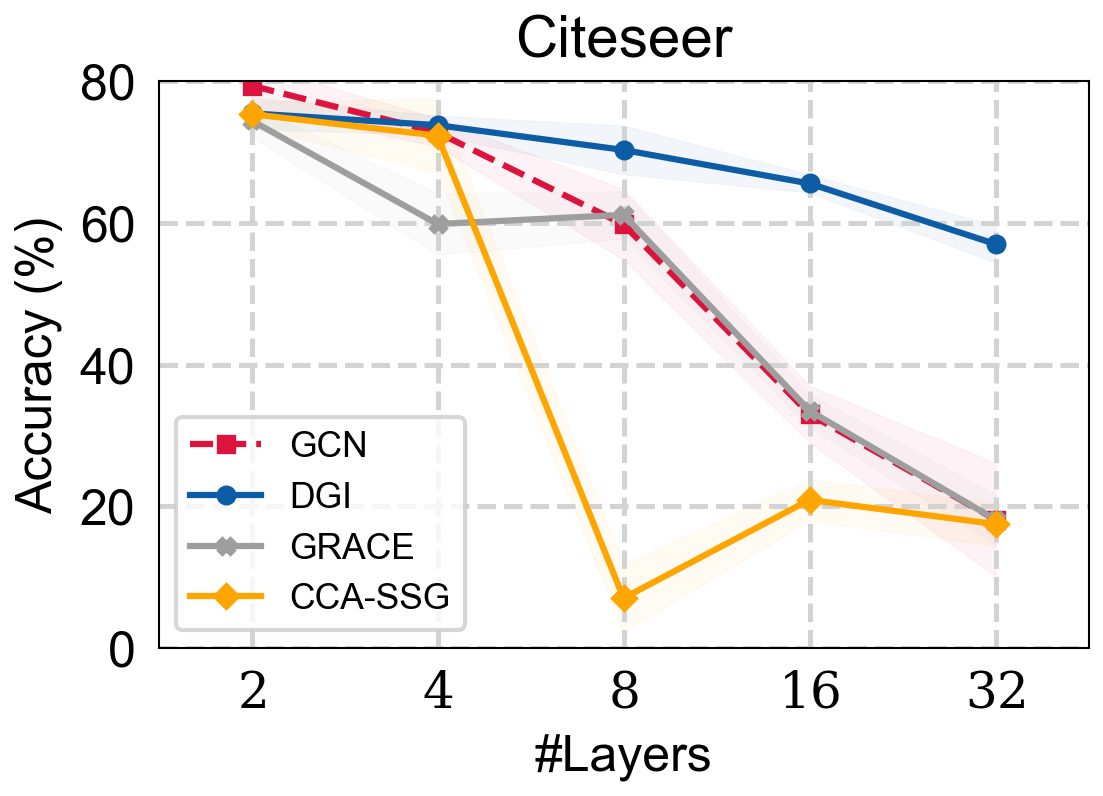}
    \caption{Visualization for the classification accuracy of GCN and three GCL methods w.r.t. the increasing layers/depth. All of the GCL methods use GCN as the encoder network.}
    \label{fig:obs1}
\end{figure}

The classification results of different methods with increasing depth are shown in Figure~\ref{fig:obs1}. We can observe that GCL methods suffer from the same oversmoothing problems as GCN, in which the oversmoothing issue becomes more severe while the number of layers grows larger.
Particularly, the performance of GCL methods drops significantly with increasing network depth. This draws our attention to the oversmoothing problem in GCL.

\subsection{Long-range starvation in GCL}
\label{sec:empirical_2}
\begin{figure}
    \centering
    \includegraphics[width=0.6\linewidth]{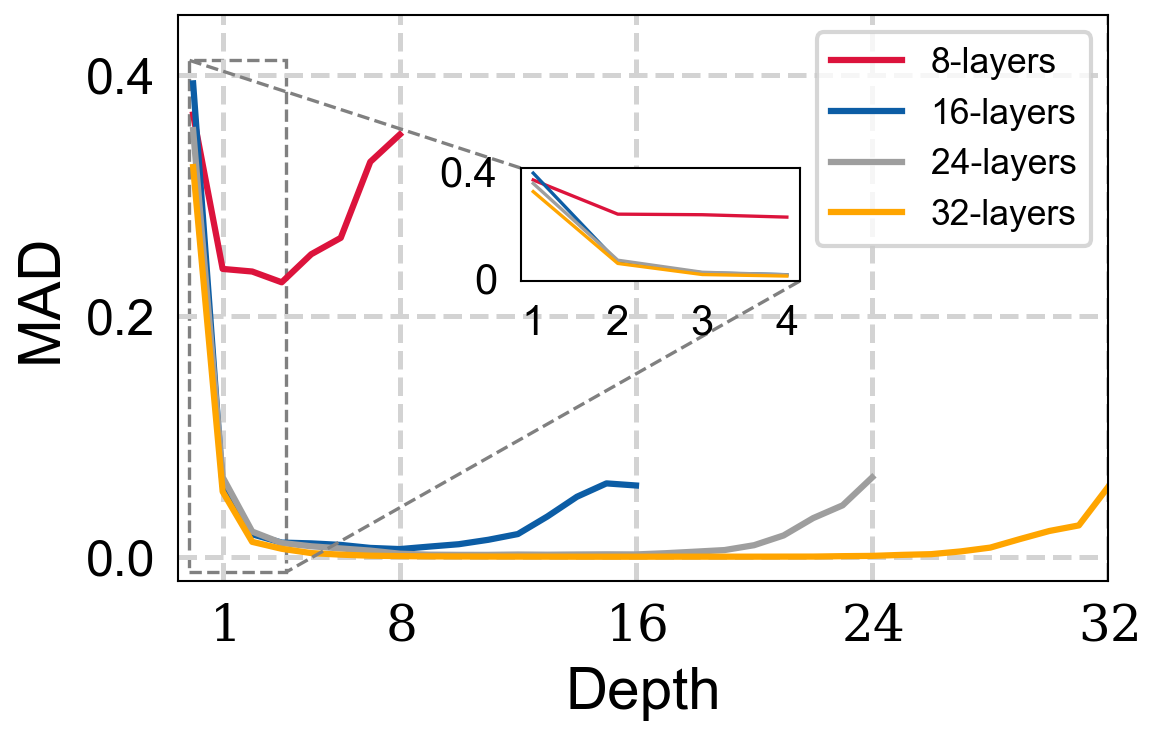}
    \caption{The MAD value of CCA-SSG with different depths. A smaller MAD value indicates a more significant oversmoothing phenomenon.}
    \label{fig:obs2}
\end{figure}
We further explore the oversmoothing issue in GCL by looking into the smoothness of node representations at the output of each layer.
The smoothness of node representations is measured by the MAD metric~\cite{mad}, which is defined as the mean of the average cosine distance from connected nodes. A smaller MAD value indicates a more significant oversmoothing phenomenon.
We follow the same experimental setting as in Section~\ref{sec:empirical_1} and report the MAD value of node representations learned by CCA-SSG with 8, 16, 24, and 32 layers, respectively. The result of CCA-SSG on Cora dataset is shown in Figure~\ref{fig:obs2}.

From Figure~\ref{fig:obs2}, we can see that increasing layers in CCA-SSG leads to node representations that collapse to be equivalent over the entire graph. Furthermore, we have observed an interesting phenomenon where shallow representations (with a depth ranging from 2 to 4) also collapse to a smoothing situation, with a significant drop beginning from the first layer. Surprisingly, we have also observed an increase in the MAD values of the representations at the last few layers for all cases, indicating mitigation of the oversmoothing phenomenon.
Given that the contrastive loss is typically performed on the last output representation, we attribute the problem to \textit{`long-range starvation'} and hypothesize that lower layers that do not receive enough supervision may lead to a suboptimal solution and fail to contribute to the overall performance of the model.
Therefore, addressing the problem of long-range starvation is an important step toward achieving better performance in deep GCL models.

The above findings raise the need to investigate better mechanisms for training deep GCL methods to prevent serious degradation in lower layers. In what follows, we present our work on mitigating the long-range starvation problem and taking one step further to address the oversmoothing problem through a new local training paradigm.

\section{Present work}
\begin{figure}[t]
    \centering
    \includegraphics[width=\linewidth]{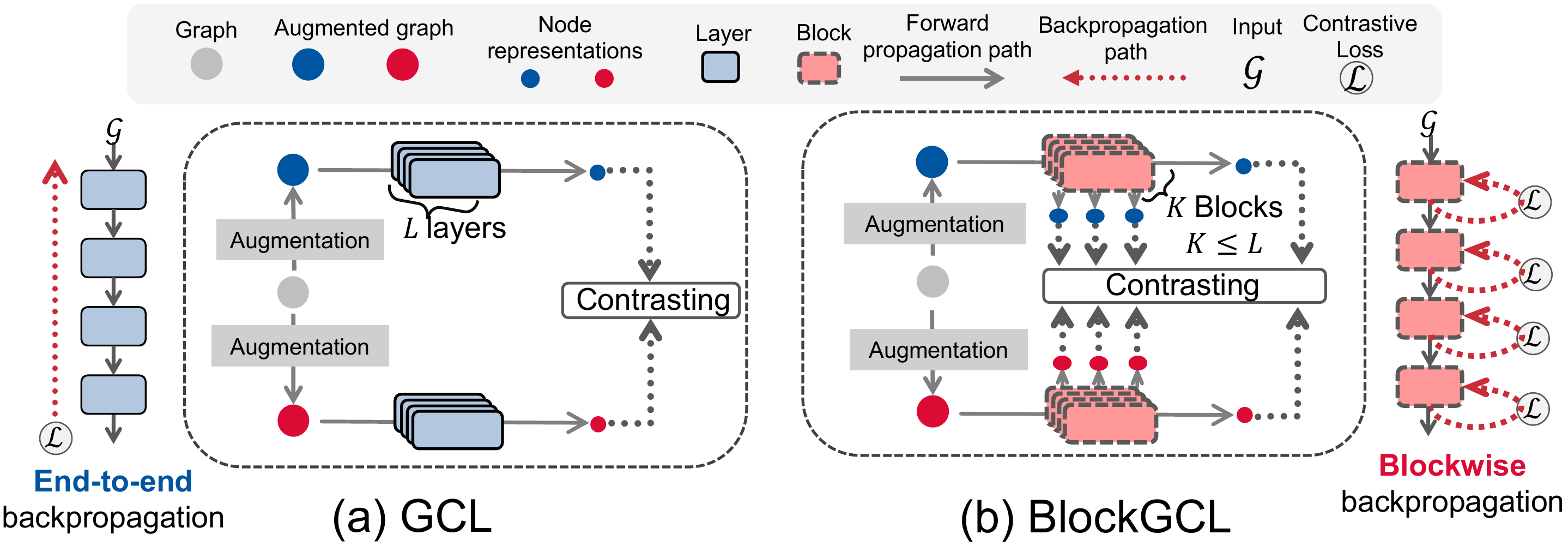}
    \caption{Technical comparison between (a) conventional GCL framework and (b) our proposed \ours framework. \ours divides networks into several non-overlapping blocks, where each block is explicitly and locally guided by a contrastive loss. In this example, the block size is set as 1 --- each block contains one layer.}
    \label{fig:framework}
\end{figure}

The key contribution of this paper is a novel blockwise training paradigm for GCL, called \ours.
As shown in Figure~\ref{fig:framework}, \ours is conceptually different from conventional GCL. Specifically, \ours divides the encoder network into several non-overlapping blocks, in which each block consists of one or more consecutive layers. Then, each block is guided by an individual GCL loss, coupled with \textit{stop-gradient} to ensure that the updates of lower layers do not directly depend on the computation of upper layers.
Here we use GCN as the encoder and adapt the CCA-SSG architecture~\cite{cca_ssg} to a blockwise training paradigm to implement \ours.

\subsection{Graph convolutional network (GCN)}
As a classical architecture of GNNs, GCN~\cite{kipf2016semi} has been widely adopted as an encoder network in many GCL methods~\cite{dgi,mvgrl,ggd,cca_ssg} and has shown remarkable potential of exploring underlying graph structure. In each round, GCN recursively and progressively propagates messages between 1-hop neighbors for exchanging information.
The message passing scheme of GCN at each layer $l$ is:
\begin{equation}
    h_u^{(l)}=\relu\left(\sum_{v \in \mathcal{N}_u \cup\{u\}} \frac{\mathbf{W}^{(l)} h_{v}^{(l-1)}}{\sqrt{\left(\left|\mathcal{N}_u\right|+1\right) \left(\mid \mathcal{N}_{v }|+1\right)}} \right)
\end{equation}
where $\mathbf{W}^{(l)}$ is the learnable parameters at $l$-th layer.

\subsection{The CCA-SSG contrastive architecture}
CCA-SSG~\cite{cca_ssg} borrows the idea of classical Canonical Correlation Analysis to self-supervised graph learning, which aims at correlating the representations of two views from data augmentation as well as decorrelating different feature dimensions of each view's representation. Formally, the CCA-SSG loss is defined as follows:
\begin{equation}
    \mathcal{L}=\underbrace{\left\|\tilde{\mathbf{Z}}_A-\tilde{\mathbf{Z}}_B\right\|_F^2}_{\text {invariance term }}+\lambda \underbrace{\left(\left\|\tilde{\mathbf{Z}}_A^{\top} \tilde{\mathbf{Z}}_A-\mathbf{I}\right\|_F^2+\left\|\tilde{\mathbf{Z}}_B^{\top} \tilde{\mathbf{Z}}_B-\mathbf{I}\right\|_F^2\right)}_{\text {decorrelation term }},
\end{equation}
where $\lambda$ is the trade-off of two terms.
$\mathbf{I}$ is an identity matrix. $\tilde{\mathbf{Z}}_A$ and $\tilde{\mathbf{Z}}_B$ are normalized representations derived from two graph augmentation views $\mathcal{G}_A$ and $\mathcal{G}_B$, respectively. Normalization is applied along the instance dimension to adjust the mean to 0 and the standard deviation to $1/\sqrt{N}$:
\begin{equation}
    \tilde{\mathbf{Z}}=\frac{\mathbf{Z}-\mu(\mathbf{Z})}{\sigma(\mathbf{Z}) * \sqrt{N}},
\end{equation}
where $\mu(\mathbf{Z})$ and $\sigma(\mathbf{Z})$ are mean and standard deviation of $\mathbf{Z}$, respectively. Despite most of our investigation being restricted to CCA-SSG, our findings and solution generalize to other GCL methods such as BGRL~\cite{bgrl}, DGI~\cite{dgi}, and GGD~\cite{ggd}, as shown in Section~\ref{sec:other_gcl}.

\subsection{Blockwise training}
Although end-to-end backpropagation is the most successful paradigm for learning modern neural networks, it has also been criticized for the lack of biological plausibility and  training efficiency, as it requires a long-range backpropagation path with symmetric weights to the feed-forward path~\cite{blockssl}. This makes it suffer from many problems such as overfitting and vanishing gradients.
As viable alternatives of end-to-end backpropagation learning, local learning paradigms have proceeded to offer better biological plausibility and training efficiency~\cite{loco,ClarkAC21,blockssl}.

In this work, we revisit the possibility of using local learning rules as an alternative for end-to-end backpropagation, and seek to continue the success of local learning paradigms in graph contrastive learning. Our work provides a simple blockwise training framework to address the oversmoothing problem. The core design behind \ours is based on the use of isolated blocks (e.g., a single layer) in networks, which are coupled with local objectives that provide guidance for each block without relying on a full backpropagation path from top to bottom. Assume that the networks are partitioned into $K$ blocks, we formalize the blockwise training on each block $i \in \{1, \ldots, K\}$ as follows:
\begin{align}
    \textbf{Forward:}  & \quad \mathcal{L}_{i} = \text{Contrast}\left(\mathbf{Z}^{(i)}_{A}, \mathbf{Z}^{(i)}_{B}\right)                                                                            \\
    \textbf{Bcakward:} & \quad \nabla_{ \mathbf{W}^{(i)}}=\frac{\partial \mathcal{L}_{i}}{\partial \mathbf{W}^{(i)}},\quad \frac{\partial \mathcal{L}_{j}}{\partial \mathbf{W}^{(i)}}=0,\ j\neq i,
\end{align}
where $\mathbf{Z}^{(i)}$ and $\mathbf{W}^{(i)}$ are the hidden representation and learnable parameters of the $i$-th block, respectively. $\text{Contrast}(\cdot,\cdot)$ is the contrastive loss performed on the representations, such as the CCA-SSG loss.

Technically, such a learning paradigm enjoys several key advantages: (i)  Bottom layers are not necessary to `wait' for upper layers. All the blocks are explicitly guided to produce discriminative representations against oversmoothing, which speeds up training and promotes convergence as well. (ii) The backpropagation path is limited within each block, potentially avoiding several notorious training problems such as overfitting or vanishing gradients. (iii) The blockwise training paradigm does not necessarily introduce additional parameters or learning costs, while enabling the training of deep networks with lower memory footprints.
To help better understand the proposed \ours, we provide the PyTorch style pseudocode for blockwise training in Algorithm~\ref{algo:pseudo_blockgcl}.

\begin{algorithm}[t]
    \caption{PyTorch style pseudocode for \ours}
    \label{algo:pseudo_blockgcl}
    \begin{algorithmic}[0]
        \State \PyComment{blocks:\ blocks of encoder network}
        \State \PyComment{g:\ input graph}
        \State \PyComment{x:\ node features}
        \State
        \State \PyComment{graph augmentation}
        \State \PyCode{x1, g1 = augmentation(x, g)}
        \State \PyCode{x2, g2 = augmentation(x, g)}
        \State \PyComment{blockwise training}
        \State \PyCode{for block in blocks:}
        \State \quad\quad\PyComment{gradient isolation}
        \State \PyCode{\quad\quad x1 = block(x1.\textcolor{purple}{detach())}, g1)}
        \State \PyCode{\quad\quad x2 = block(x2.\textcolor{purple}{detach())}, g2)}
        \State \PyCode{\quad\quad loss = contrastive\_loss(x1, x2)}
        \State \PyCode{\quad\quad loss.\textcolor{purple}{backward()}}
    \end{algorithmic}
\end{algorithm}

\section{Experiments}
\label{sec:exp}

In this section, we conducted extensive experimental evaluations on several real-world datasets to demonstrate the effectiveness of \ours.
In what follows, we first introduce the experimental settings and then present empirical results.

\subsection{Experimental setup}

\nosection{Datasets.}
We evaluate \ours on several graph benchmarks, including three citation graphs Cora, Citeseer, Pubmed~\cite{sen2008collective}, and two co-purchase graphs Amazon-Photo, Amazon-Computer~\cite{shchur2018pitfalls}.
We follow the experimental public splits stated in \cite{dropedge}. For datasets without public splits, we follow the recommended splits in \cite{shchur2018pitfalls}, which adopt a random 1:1:8 split for Amazon-Photo and Amazon-Computer datasets.
Dataset statistics are listed in Table~\ref{tab:dataset}.

\nosection{Comparison methods.}
We compare our proposed methods to a series of supervised and self-supervised contrastive methods: (i) supervised GNNs: GCN~\cite{kipf2016semi} and GAT~\cite{gat}; (ii) self-supervised contrastive GNNs: DGI~\cite{dgi}, GRACE~\cite{grace}, GGD~\cite{ggd}, SUGRL~\cite{sugrl}, BGRL~\cite{bgrl}, and CCA-SSG~\cite{cca_ssg}. Among them, DGI, GRACE, GGD, and SUGRL are negative-sample-based methods while BGRL and CCA-SSG are negative-sample-free methods.
For all of the comparison methods, we basically adopt the hyperparameter settings officially reported by their respective authors, and then perform a grid search for the hyperparameters to achieve the best performance in our experiments.

\begin{table}[t]
    \centering
    \begin{threeparttable}
        \resizebox{\linewidth}{!}
        {\begin{tabular}{l|l l l l l}
                \toprule
                                    & \textbf{Cora} & \textbf{Citeseer} & \textbf{Pubmed} & \textbf{Photo} & \textbf{Computer} \\
                \midrule
                \textbf{\#Nodes}    & 2,708         & 3,327             & 19,717          & 7,650          & 13,752            \\
                \textbf{\#Edges}    & 10,556        & 9,104             & 88,648          & 238,162        & 491,722           \\
                \textbf{\#Features} & 1,433         & 3,703             & 500             & 745            & 767               \\
                \textbf{\#Classes}  & 7             & 6                 & 3               & 8              & 10                \\
                \bottomrule
            \end{tabular}
        }
    \end{threeparttable}
    \caption{Dataset statistics.}\label{tab:dataset}
\end{table}

\nosection{Implementation details.}
We exactly follow the conventional experimental setup of graph self-supervised learning from \cite{dgi} and perform a linear probing scheme evaluation. To be specific, we train a linear classifier (e.g., a logistic regression model) on top of the frozen representations and report averaged accuracy with standard deviation across 5 different runs.
\ours is built upon PyTorch~\cite{pytorch} and PyTorch Geometric~\cite{pyg}. All datasets used throughout experiments are publicly available in the PyTorch Geometric library. All experiments are done on a single NVIDIA 3090 Ti GPU (with 24GB memory).

\nosection{Hyperparameter settings.}
We follow closely the original CCA-SSG~\cite{cca_ssg} training procedure.
We set the block size as 1 across all datasets and we discuss it in Appendix.
For consistency, we also use edge dropping and feature masking for graph augmentations, where the dropping/masking ratio is tuned from $\{0.1, \ldots, 0.9\}$.

\begin{table}[t]
    \centering
    \begin{threeparttable}{
            \resizebox{\linewidth}{!}{
                \begin{tabular}{lccccc}
                    \toprule
                             & \textbf{Cora}          & \textbf{Citeseer}      & \textbf{Pubmed}        & \textbf{Photo}         & \textbf{Computer}      \\
                    \midrule
                    GCN      & 86.1$_{\pm0.2}$        & 75.9$_{\pm0.4}$        & 88.2$_{\pm0.5}$        & 92.4$_{\pm0.2}$        & 86.5$_{\pm0.5}$        \\
                    GAT      & 86.7$_{\pm0.7}$        & \Scnd{78.5$_{\pm0.4}$} & 86.8$_{\pm0.3}$        & 92.6$_{\pm0.4}$        & 86.9$_{\pm0.3}$        \\
                    \midrule
                    DGI      & 86.3$_{\pm0.2}$        & \Frst{78.9$_{\pm0.2}$} & 86.2$_{\pm0.1}$        & 91.6$_{\pm0.2}$        & 84.0$_{\pm0.5}$        \\
                    GGD      & 86.2$_{\pm0.2}$        & 75.5$_{\pm0.1}$        & 84.2$_{\pm0.1}$        & 92.9$_{\pm0.2}$        & 88.0$_{\pm0.1}$        \\
                    GRACE    & 87.2$_{\pm0.2}$        & 74.5$_{\pm0.1}$        & 87.3$_{\pm0.1}$        & 92.2$_{\pm0.2}$        & 86.3$_{\pm0.3}$        \\
                    SUGRL    & \Scnd{88.0$_{\pm0.1}$} & 77.6$_{\pm0.4}$        & 88.2$_{\pm0.2}$        & \Scnd{93.2$_{\pm0.4}$} & 88.9$_{\pm0.2}$        \\
                    BGRL     & 87.3$_{\pm0.1}$        & 76.0$_{\pm0.2}$        & 88.3$_{\pm0.1}$        & \Scnd{93.2$_{\pm0.3}$} & \Frst{90.3$_{\pm0.2}$} \\
                    CCA-SSG  & 84.6$_{\pm0.7}$        & 75.4$_{\pm1.0}$        & \Scnd{88.4$_{\pm0.6}$} & 93.1$_{\pm0.1}$        & 88.7$_{\pm0.3}$        \\
                    \midrule
                    BlockGCL & \Frst{88.1$_{\pm0.1}$} & \Frst{78.9$_{\pm0.6}$} & \Frst{88.6$_{\pm0.2}$} & \Frst{93.3$_{\pm0.1}$} & \Scnd{89.4$_{\pm0.1}$} \\
                    \bottomrule
                \end{tabular}}
        }
    \end{threeparttable}
    \caption{Performance of different methods at their optimal depths. The best and the second results of each dataset are highlighted in \Frst{bold} and \Scnd{underlined}, respectively.}\label{tab:node_clas}
\end{table}

\subsection{Comparison study}

In literature, it has been empirically demonstrated that shallow networks are sufficient for GNNs to learn good node representations in literature, and deep architectures generally do not show significant advantages over shallow ones~\cite{nodenorm,dropedge}. In this regard, we first investigate the best performance of \ours in standard cases with just a handful of layers, results are summarized in Table~\ref{tab:node_clas}. All of the comparison methods adopt their best settings in our experiments including the network depth (i.e., 1 or 2).
From Table~\ref{tab:node_clas}, we can observe that \ours can reach performance on par with advanced supervised and contrastive methods trained end-to-end with backpropagation across all datasets.
As a derivative of CCA-SSG, \ours boosts the performance of CCA-SSG and achieves state-of-the-art performance in most cases by incorporating a blockwise training paradigm.
Specifically, \ours is among the top-2 best-performing methods and achieves the best performance in 4 out of 5 cases.
Overall, our results validate the effectiveness of \ours and demonstrate that blockwise training is a promising paradigm for training GCL methods without sacrificing performance.

\begin{table}[t]
    \centering
    \resizebox{\linewidth}{!}
    {
        \begin{tabular}{llccccccc}
            \toprule
                                                               & \textbf{Depth} & \textbf{DGI}    & \textbf{GGD}    & \textbf{GRACE}  & \textbf{SUGRL}         & \textbf{BGRL}          & \textbf{CCA-SSG}       & \textbf{\ours}           \\
            \cmidrule{2-9}
            \multirow{6}{*}{\rotatebox{90}{\textbf{Cora}}}     & \#L=2          & 83.9$_{\pm0.3}$ & 83.0$_{\pm0.1}$ & 87.2$_{\pm0.2}$ & \Scnd{88.0$_{\pm0.1}$} & 87.3$_{\pm0.1}$        & 84.6$_{\pm0.7}$        & \Frst{88.1$_{\pm0.1}$}
            \\
                                                               & \#L=4          & 68.4$_{\pm0.4}$ & 77.3$_{\pm0.4}$ & 80.0$_{\pm0.3}$ & 83.7$_{\pm0.1}$        & \Scnd{85.2$_{\pm0.3}$} & 81.9$_{\pm1.3}$        & \Frst{88.0$_{\pm0.2}$}   \\
                                                               & \#L=8          & 54.7$_{\pm0.4}$ & 65.4$_{\pm0.0}$ & 75.2$_{\pm0.3}$ & 72.3$_{\pm0.1}$        & \Scnd{80.7$_{\pm0.2}$} & 64.8$_{\pm6.8}$        & \Frst{86.6$_{\pm0.1}$}   \\
                                                               & \#L=16         & 62.1$_{\pm0.3}$ & 45.2$_{\pm0.5}$ & 61.6$_{\pm0.3}$ & 69.0$_{\pm0.1}$        & \Scnd{76.4$_{\pm0.7}$} & 16.8$_{\pm3.9}$        & \Frst{84.6$_{\pm0.4}$}   \\
                                                               & \#L=32         & 55.1$_{\pm0.6}$ & 40.4$_{\pm2.4}$ & 31.9$_{\pm0.0}$ & 31.9$_{\pm0.0}$        & \Scnd{58.4$_{\pm0.3}$} & 18.8$_{\pm3.1}$        & \Frst{82.9$_{\pm0.5}$}   \\
            \midrule

                                                               & \textbf{Depth} & \textbf{DGI}    & \textbf{GGD}    & \textbf{GRACE}  & \textbf{SUGRL}         & \textbf{BGRL}          & \textbf{CCA-SSG}       & \textbf{\ours}           \\
            \cmidrule{2-9}
            \multirow{6}{*}{\rotatebox{90}{\textbf{Citeseer}}} & \#L=2          & 75.5$_{\pm0.2}$ & 73.2$_{\pm0.1}$ & 74.5$_{\pm0.1}$ & 74.3$_{\pm0.2}$        & \Scnd{76.0$_{\pm0.2}$} & 75.4$_{\pm1.0}$        & \Frst{78.9$_{\pm0.6}$}   \\
                                                               & \#L=4          & 73.8$_{\pm0.1}$ & 71.2$_{\pm0.1}$ & 59.9$_{\pm0.2}$ & 71.9$_{\pm0.2}$        & \Scnd{73.6$_{\pm0.5}$} & 72.3$_{\pm1.6}$        & \Frst{76.4$_{\pm1.2}$}   \\
                                                               & \#L=8          & 70.3$_{\pm0.2}$ & 47.7$_{\pm0.3}$ & 61.2$_{\pm0.2}$ & 33.0$_{\pm0.1}$        & \Scnd{64.6$_{\pm0.2}$} & 27.0$_{\pm0.0}$        & \Frst{75.4$_{\pm0.4}$}   \\
                                                               & \#L=16         & 65.6$_{\pm0.2}$ & 41.7$_{\pm0.6}$ & 33.5$_{\pm0.3}$ & 18.1$_{\pm0.0}$        & \Scnd{55.9$_{\pm0.1}$} & 20.9$_{\pm0.5}$        & \Frst{71.3$_{\pm0.9}$}   \\
                                                               & \#L=32         & 57.0$_{\pm0.2}$ & 30.0$_{\pm0.6}$ & 18.1$_{\pm0.0}$ & 18.1$_{\pm0.0}$        & \Scnd{50.8$_{\pm0.2}$} & 17.5$_{\pm4.4}$        & \Frst{63.1$_{\pm0.5}$}   \\
            \midrule
                                                               & \textbf{Depth} & \textbf{DGI}    & \textbf{GGD}    & \textbf{GRACE}  & \textbf{SUGRL}         & \textbf{BGRL}          & \textbf{CCA-SSG}       & \textbf{\ours}           \\
            \cmidrule{2-9}
            \multirow{6}{*}{\rotatebox{90}{\textbf{Pubmed}}}   & \#L=2          & 83.2$_{\pm0.1}$ & 84.2$_{\pm0.1}$ & 87.3$_{\pm0.1}$ & 88.2$_{\pm0.1}$        & 88.3$_{\pm0.1}$        & \Scnd{88.4$_{\pm0.6}$} & \Frst{88.6$_{\pm0.2}$}   \\
                                                               & \#L=4          & 82.4$_{\pm0.1}$ & 81.0$_{\pm0.0}$ & 84.2$_{\pm0.6}$ & 84.5$_{\pm0.1}$        & \Frst{86.3$_{\pm0.2}$} & 84.7$_{\pm0.8}$        & \Scnd{85.6$_{\pm0.2}$}   \\
                                                               & \#L=8          & 70.0$_{\pm0.1}$ & 69.5$_{\pm0.1}$ & 81.2$_{\pm0.2}$ & 69.9$_{\pm0.0}$        & \Scnd{83.9$_{\pm0.1}$} & 71.0$_{\pm4.7}$        & \Frst{84.1$_{\pm0.1}$}   \\
                                                               & \#L=16         & 68.5$_{\pm0.4}$ & 58.9$_{\pm0.7}$ & 67.9$_{\pm0.2}$ & 40.7$_{\pm0.0}$        & \Scnd{82.1$_{\pm0.3}$} & 52.6$_{\pm4.0}$        & \Frst{82.6$_{\pm0.4}$}   \\
                                                               & \#L=32         & OOM             & 44.0$_{\pm2.3}$ & 45.8$_{\pm0.9}$ & 40.7$_{\pm0.0}$        & \Scnd{78.4$_{\pm0.3}$} & 40.5$_{\pm3.6}$        & \Frst{80.5$_{\pm0.3}$}   \\
            \midrule
                                                               & \textbf{Depth} & \textbf{DGI}    & \textbf{GGD}    & \textbf{GRACE}  & \textbf{SUGRL}         & \textbf{BGRL}          & \textbf{CCA-SSG}       & \textbf{\ours}           \\
            \cmidrule{2-9}
            \multirow{6}{*}{\rotatebox{90}{\textbf{Photo}}}    & \#L=2          & 90.9$_{\pm0.0}$ & 92.4$_{\pm0.1}$ & 92.1$_{\pm0.0}$ & 91.9$_{\pm0.0}$        & \Scnd{93.2$_{\pm0.2}$} & 93.1$_{\pm0.1}$        & \Frst{93.3$_{\pm0.1}$}   \\
                                                               & \#L=4          & 88.1$_{\pm0.1}$ & 90.6$_{\pm0.0}$ & 91.1$_{\pm0.1}$ & 89.4$_{\pm0.1}$        & \Scnd{91.9$_{\pm0.3}$} & 90.2$_{\pm0.5}$        & \Frst{92.0$_{\pm0.1}$}   \\
                                                               & \#L=8          & 84.3$_{\pm0.1}$ & 86.6$_{\pm0.2}$ & 88.7$_{\pm0.1}$ & 84.3$_{\pm0.0}$        & \Scnd{90.0$_{\pm0.2}$} & 86.3$_{\pm0.9}$        & \Frst{90.7$_{\pm0.0}$}   \\
                                                               & \#L=16         & 71.1$_{\pm1.4}$ & 67.2$_{\pm0.6}$ & 66.9$_{\pm0.3}$ & 25.3$_{\pm0.1}$        & \Scnd{87.9$_{\pm0.4}$} & 48.5$_{\pm12.}$        & \Frst{89.4$_{\pm0.0}$}   \\
                                                               & \#L=32         & 62.8$_{\pm0.0}$ & 39.1$_{\pm0.0}$ & 58.1$_{\pm0.2}$ & 25.3$_{\pm0.0}$        & \Scnd{84.3$_{\pm0.3}$} & 38.1$_{\pm2.6}$        & \Frst{87.5$_{\pm0.1}$}   \\
            \midrule
                                                               & \textbf{Depth} & \textbf{DGI}    & \textbf{GGD}    & \textbf{GRACE}  & \textbf{SUGRL}         & \textbf{BGRL}          & \textbf{CCA-SSG}       & \textbf{\ours}           \\
            \cmidrule{2-9}
            \multirow{6}{*}{\rotatebox{90}{\textbf{Computer}}} & \#L=2          & 82.9$_{\pm0.0}$ & 87.2$_{\pm0.1}$ & 86.3$_{\pm0.3}$ & 87.0$_{\pm0.1}$        & \Frst{90.3$_{\pm0.2}$} & 88.7$_{\pm0.3}$        & \Scnd{89.4$_{\pm0.1}$} \
            \\
                                                               & \#L=4          & 76.0$_{\pm0.1}$ & 82.8$_{\pm0.2}$ & 84.8$_{\pm0.0}$ & 84.1$_{\pm0.3}$        & \Scnd{88.1$_{\pm0.3}$} & 82.2$_{\pm0.5}$        & \Frst{88.6$_{\pm0.1}$}   \\
                                                               & \#L=8          & 71.1$_{\pm0.0}$ & 75.8$_{\pm0.0}$ & 80.1$_{\pm0.2}$ & 78.1$_{\pm0.3}$        & \Frst{87.2$_{\pm0.4}$} & 73.9$_{\pm0.9}$        & \Scnd{87.1$_{\pm0.0}$}   \\
                                                               & \#L=16         & 51.8$_{\pm0.5}$ & 56.4$_{\pm0.2}$ & 67.3$_{\pm0.0}$ & 37.5$_{\pm0.5}$        & \Scnd{85.1$_{\pm0.4}$} & 39.1$_{\pm1.2}$        & \Frst{85.6$_{\pm0.1}$}   \\
                                                               & \#L=32         & 45.8$_{\pm2.1}$ & 38.1$_{\pm0.3}$ & 69.4$_{\pm0.4}$ & 37.5$_{\pm0.2}$        & \Scnd{84.3$_{\pm0.4}$} & 39.1$_{\pm1.4}$        & \Frst{84.7$_{\pm0.1}$}   \\

            \bottomrule
        \end{tabular}
    }
    \caption{Performance of different GCL methods w.r.t. increasing network depth. The best and the second results in each row are highlighted in \Frst{bold} and \Scnd{underlined}, respectively. (OOM: out-of-memory on a GPU with 24GB memory)}\label{tab:deep}
\end{table}

\subsection{Robustness against oversmoothing}

To further investigate whether \ours can help resolve performance degradation of GCL using deep architectures, we conduct experiments on the datasets with varying numbers of layers/depth $L \in \{2, 4, 8, 16, 32\}$. The results of different GCL methods are shown in Table~\ref{tab:deep}.

As observed from Table~\ref{tab:deep}, most GCL methods suffer from severe oversmoothing issues with increasing network depth, especially GRACE, SUGRL, and CCA-SSG. It is obvious that the learned representations by these methods are collapsed, leading to poor performance with a network depth significantly larger than $8$.
In contrast, \ours exhibits remarkable robustness against oversmoothing, which easily scales to deeper layers without significantly sacrificing performance.
In particular, \ours achieves state-of-the-art performance across most cases.
With the increasing number of layers, \ours outperforms other baselines results by larger margins.
By incorporating a simple blockwise training design, \ours successfully closes the performance gap between deep and shallow contrastive learning algorithms for the first time.
Overall, the results demonstrate that \ours achieves promising performance in alleviating the oversmoothing issue.

\begin{table}[t]
    \centering
    \resizebox{\linewidth}{!}
    {
        \begin{tabular}{l l ccccc}
            \toprule
             &        & \#L=2                            & \#L=4                            & \#L=8                            & \#L=16                           & \#L=32                           \\
            \midrule
            \multirow{6}{*}{\textbf{Cora}}
             & GRACE  & 87.2$_{\pm0.2}$                  & 80.0$_{\pm0.3}$                  & 75.2$_{\pm0.3}$                  & 61.6$_{\pm0.3}$                  & 31.9$_{\pm0.0}$                  \\
             & +\ours & {\colorbox{c1}{88.1$_{\pm0.2}$}} & {\colorbox{c1}{84.7$_{\pm0.2}$}} & {\colorbox{c1}{79.2$_{\pm0.4}$}} & {\colorbox{c3}{78.0$_{\pm0.3}$}} & {\colorbox{c5}{71.6$_{\pm0.1}$}} \\
             & SUGRL  & 88.0$_{\pm0.1}$                  & 83.7$_{\pm0.1}$                  & 72.3$_{\pm0.1}$                  & 69.0$_{\pm0.1}$                  & 31.9$_{\pm0.0}$                  \\
             & +\ours & 87.0$_{\pm0.3}$                  & {\colorbox{c1}{85.6$_{\pm0.2}$}} & {\colorbox{c2}{85.0$_{\pm0.2}$}} & {\colorbox{c3}{83.4$_{\pm0.2}$}} & {\colorbox{c5}{76.1$_{\pm0.1}$}} \\
             & BGRL   & 87.3$_{\pm0.1}$                  & 85.2$_{\pm0.3}$                  & 80.7$_{\pm0.2}$                  & 76.4$_{\pm0.7}$                  & 58.4$_{\pm0.3}$                  \\
             & +\ours & {\colorbox{c1}{87.4$_{\pm0.3}$}} & {\colorbox{c1}{86.9$_{\pm0.4}$}} & {\colorbox{c1}{84.4$_{\pm0.2}$}} & {\colorbox{c1}{81.5$_{\pm0.1}$}} & {\colorbox{c3}{78.0$_{\pm0.3}$}} \\

            \midrule
            \multirow{6}{*}{\textbf{Citeseer}}
             & GRACE  & {74.5$_{\pm0.1}$}                & 59.9$_{\pm0.2}$                  & 61.2$_{\pm0.2}$                  & 33.5$_{\pm0.3}$                  & 18.1$_{\pm0.0}$                  \\
             & +\ours & 74.5$_{\pm0.3}$                  & {\colorbox{c2}{74.4$_{\pm0.1}$}} & {\colorbox{c2}{72.3$_{\pm0.2}$}} & {\colorbox{c3}{67.8$_{\pm0.5}$}} & {\colorbox{c4}{53.0$_{\pm2.3}$}} \\
             & SUGRL  & 74.3$_{\pm0.2}$                  & 72.0$_{\pm0.2}$                  & 33.0$_{\pm0.1}$                  & 18.1$_{\pm0.0}$                  & 18.1$_{\pm0.0}$                  \\
             & +\ours & {\colorbox{c1}{76.7$_{\pm0.3}$}} & {\colorbox{c1}{75.7$_{\pm0.2}$}} & {\colorbox{c4}{73.3$_{\pm0.2}$}} & {\colorbox{c5}{68.1$_{\pm0.1}$}} & {\colorbox{c4}{53.2$_{\pm0.3}$}} \\
             & BGRL   & 76.0$_{\pm0.2}$                  & 73.6$_{\pm0.5}$                  & 64.6$_{\pm0.2}$                  & 55.9$_{\pm0.1}$                  & 50.8$_{\pm0.2}$                  \\
             & +\ours & {\colorbox{c1}{76.4$_{\pm0.2}$}} & {\colorbox{c1}{75.6$_{\pm1.0}$}} & {\colorbox{c2}{74.5$_{\pm0.1}$}} & {\colorbox{c3}{73.2$_{\pm0.5}$}} & {\colorbox{c3}{72.0$_{\pm0.4}$}} \\

            \midrule
            \multirow{6}{*}{\textbf{Pubmed}}
             & GRACE  & {87.3$_{\pm0.1}$}                & 84.2$_{\pm0.6}$                  & 81.2$_{\pm0.2}$                  & 67.9$_{\pm0.2}$                  & 45.8$_{\pm0.9}$                  \\
             & +\ours & 87.0$_{\pm0.3}$                  & {\colorbox{c1}{84.2$_{\pm0.2}$}} & {\colorbox{c1}{82.5$_{\pm0.2}$}} & {\colorbox{c2}{78.4$_{\pm0.3}$}} & {\colorbox{c4}{73.9$_{\pm0.7}$}} \\
             & SUGRL  & 88.2$_{\pm0.1}$                  & 84.5$_{\pm0.1}$                  & 69.9$_{\pm0.0}$                  & 40.7$_{\pm0.0}$                  & 40.7$_{\pm0.0}$                  \\
             & +\ours & 88.1$_{\pm0.1}$                  & {\colorbox{c1}{86.1$_{\pm0.2}$}} & {\colorbox{c2}{83.9$_{\pm0.1}$}} & {\colorbox{c4}{73.1$_{\pm0.1}$}} & {\colorbox{c3}{60.2$_{\pm0.2}$}} \\
             & BGRL   & 88.3$_{\pm0.1}$                  & {86.3$_{\pm0.2}$}                & {83.9$_{\pm0.1}$}                & {82.1$_{\pm0.3}$}                & 78.4$_{\pm0.3}$                  \\
             & +\ours & {88.3$_{\pm0.1}$}                & 86.0$_{\pm0.2}$                  & 83.9$_{\pm0.2}$                  & 81.7$_{\pm0.1}$                  & {\colorbox{c1}{80.7$_{\pm0.2}$}} \\

            \bottomrule
        \end{tabular}
    }
    \caption{Performance of different GCL methods coupled with \ours. {\colorbox{c4} {Darker color}} means larger improvements over baseline methods.}
    \label{tab:block}
\end{table}

\subsection{Applying \ours to other GCL methods}
\label{sec:other_gcl}

To further verify the de-oversmoothing effect of \ours from a broad perspective, we apply \ours to three popular GCL architectures that also suffer seriously from performance degradation, i.e., GRACE, SUGRL, and BGRL, to study whether blockwise training can improve their robustness against oversmoothing. As shown in Table~\ref{tab:block}, \ours generalizes well to these methods, significantly mitigating their performance degradation.
In particular, \ours helps deep GCL methods to compete with their shallow counterparts and scale to deeper networks without compromising the performance. The results well demonstrate that \ours also addresses performance degradation for other GCL methods in addition to CCA-SSG.

We also note that vanilla BGRL, a negative-sample-free method, exhibits high robustness against oversmoothing than negative-sampled-based methods GRACE and SUGRL.
We conjecture that oversmoothing causes the node representations to become indistinguishable even after different augmentations, making it difficult to provide high-quality negative samples for negative-sampled-based contrastive learning. In contrast, BGRL does not rely on negative samples and has demonstrated superior robustness against oversmoothing. Therefore, \ours becomes less effective for improving the BGRL's robustness against oversmoothing in most cases.

\subsection{Comparison with de-oversmoothing techniques}
In literature, several techniques have been proposed to address the oversmoothing effect in graph neural networks. Many of these techniques are out-of-box and can be directly adapted into existing GCL frameworks without additional efforts. In this study, we compare our proposed \ours with these de-oversmoothing techniques to validate its effectiveness. The compared techniques include DropEdge~\cite{dropedge}, PairNorm~\cite{pairnorm}, NodeNorm~\cite{nodenorm}, and ContraNorm~\cite{contranorm}. All the methods are configured with their optimal settings for fair comparison.

As shown in Table~\ref{tab:deoversmoothing}, incorporating different techniques helps prevent the performance of vanilla CCA-SSG from sharply deteriorating as the number of layers increases. Among the comparison methods, DropEdge shows marginal or even negative improvements over vanilla CCA-SSG. This is because random edge dropping is already used as a graph augmentation technique in CCA-SSG. Therefore, incorporating DropEdge additionally does not result in significant improvements and can even amplify the negative effect, especially with more than 8 layers.
Furthermore, training deep GCLs with normalization can indeed enhance robustness against oversmoothing. This is consistent with the results observed in general GNNs~\cite{dropedge,pairnorm,nodenorm}.
However, they become less effective and incur more instability (e.g., large performance variance) along with the increasing depth of networks.
In general, we demonstrate that the proposed \ours is the key solution to alleviate oversmoothing, which significantly outperforms the baselines in terms of de-oversmoothing effectiveness and provides better stability.

\begin{table}[t]
    \centering
    \resizebox{\linewidth}{!}
    {
        \begin{tabular}{l l ccccc}
            \toprule
             &             & \#L=2                  & \#L=4                  & \#L=8                  & \#L=16                  & \#L=32                 \\
            \midrule
            \multirow{6}{*}{\textbf{Cora}}
             & CCA-SSG     & \Scnd{84.6$_{\pm0.7}$} & 81.9$_{\pm1.3}$        & 64.8$_{\pm6.8}$        & 16.8$_{\pm3.9}$         & 18.8$_{\pm3.1}$        \\
             & +DropEdge   & 84.3$_{\pm0.5}$        & 82.5$_{\pm0.9}$        & 67.1$_{\pm2.7}$        & 15.2$_{\pm5.1}$         & 14.8$_{\pm3.5}$        \\
             & +PairNorm   & 81.2$_{\pm1.2}$        & 79.5$_{\pm1.5}$        & 71.2$_{\pm4.3}$        & 23.5$_{\pm5.5}$         & 21.3$_{\pm6.5}$        \\
             & +NodeNorm   & 82.3$_{\pm1.5}$        & 80.5$_{\pm2.7}$        & 68.7$_{\pm3.5}$        & 17.9$_{\pm5.7}$         & 15.4$_{\pm4.6}$        \\
             & +ContraNorm & 83.9$_{\pm1.3}$        & \Scnd{84.2$_{\pm1.8}$} & \Scnd{75.1$_{\pm4.1}$} & \Scnd{33.8 $_{\pm5.2}$} & \Scnd{29.7$_{\pm5.3}$} \\
             & +\ours      & \Frst{88.1$_{\pm0.1}$} & \Frst{88.0$_{\pm0.2}$} & \Frst{86.6$_{\pm0.1}$} & \Frst{84.6$_{\pm0.4}$}  & \Frst{82.9$_{\pm0.5}$} \\
            \midrule
            \multirow{6}{*}{\textbf{Citeseer}}
             & CCA-SSG     & \Scnd{75.4$_{\pm1.0}$} & \Scnd{72.3$_{\pm1.6}$} & 27.0$_{\pm0.0}$        & 20.9$_{\pm0.5}$         & 17.5$_{\pm4.4}$        \\
             & +DropEdge   & 73.1$_{\pm0.9}$        & 70.9$_{\pm2.8}$        & 30.2$_{\pm1.3}$        & 22.7$_{\pm1.2}$         & 15.5$_{\pm4.1}$        \\
             & +PairNorm   & 68.3$_{\pm1.3}$        & 63.2$_{\pm2.4}$        & 35.5$_{\pm1.7}$        & 28.4$_{\pm2.6}$         & 28.1$_{\pm5.2}$        \\
             & +NodeNorm   & 71.4$_{\pm1.6}$        & 68.5$_{\pm1.9}$        & 33.1$_{\pm1.8}$        & 27.9$_{\pm4.6}$         & 23.5$_{\pm5.0}$        \\
             & +ContraNorm & 73.6$_{\pm1.8}$        & 65.7$_{\pm2.5}$        & \Scnd{40.1$_{\pm1.3}$} & \Scnd{35.8$_{\pm2.2}$}  & \Scnd{32.4$_{\pm6.2}$} \\
             & +\ours      & \Frst{78.9$_{\pm0.1}$} & \Frst{76.4$_{\pm1.2}$} & \Frst{75.4$_{\pm0.4}$} & \Frst{71.3$_{\pm0.9}$}  & \Frst{63.1$_{\pm0.5}$} \\

            \midrule
            \multirow{6}{*}{\textbf{Pubmed}}
             & CCA-SSG     & 88.4$_{\pm0.6}$        & 84.7$_{\pm0.8}$        & 71.0$_{\pm4.7}$        & 52.6$_{\pm4.0}$         & 40.5$_{\pm3.6}$        \\
             & +DropEdge   & \Frst{88.9$_{\pm0.4}$} & 83.7$_{\pm0.7}$        & 70.2$_{\pm5.1}$        & 47.5$_{\pm3.7}$         & 41.7$_{\pm4.5}$        \\
             & +PairNorm   & 87.5$_{\pm2.7}$        & 83.7$_{\pm1.9}$        & 72.3$_{\pm4.9}$        & 59.9$_{\pm3.4}$         & 50.1$_{\pm4.3}$        \\
             & +NodeNorm   & 87.3$_{\pm1.7}$        & 84.5$_{\pm3.4}$        & 74.3$_{\pm4.1}$        & 40.9$_{\pm4.3}$         & 31.2$_{\pm4.6}$        \\
             & +ContraNorm & 88.0$_{\pm2.6}$        & \Scnd{85.4$_{\pm2.7}$} & \Scnd{76.2$_{\pm6.3}$} & \Scnd{63.1$_{\pm5.2}$}  & \Scnd{55.7$_{\pm5.1}$} \\
             & +\ours      & \Scnd{88.6$_{\pm0.2}$} & \Frst{85.6$_{\pm0.2}$} & \Frst{84.1$_{\pm0.1}$} & \Frst{82.6$_{\pm0.4}$}  & \Frst{80.5$_{\pm0.3}$} \\

            \bottomrule
        \end{tabular}
    }
    \caption{Comparison results of different de-oversmoothing techniques. We use CCA-SSG as baseline and apply DropEdge, PairNorm, NodeNorm, ContraNorm and \ours, respectively. The best and the second results in each column are highlighted in \Frst{bold} and \Scnd{underlined}, respectively.}
    \label{tab:deoversmoothing}
\end{table}

\subsection{Convergence}
\begin{figure}[t]
    \centering
    \includegraphics[width=0.7\linewidth]{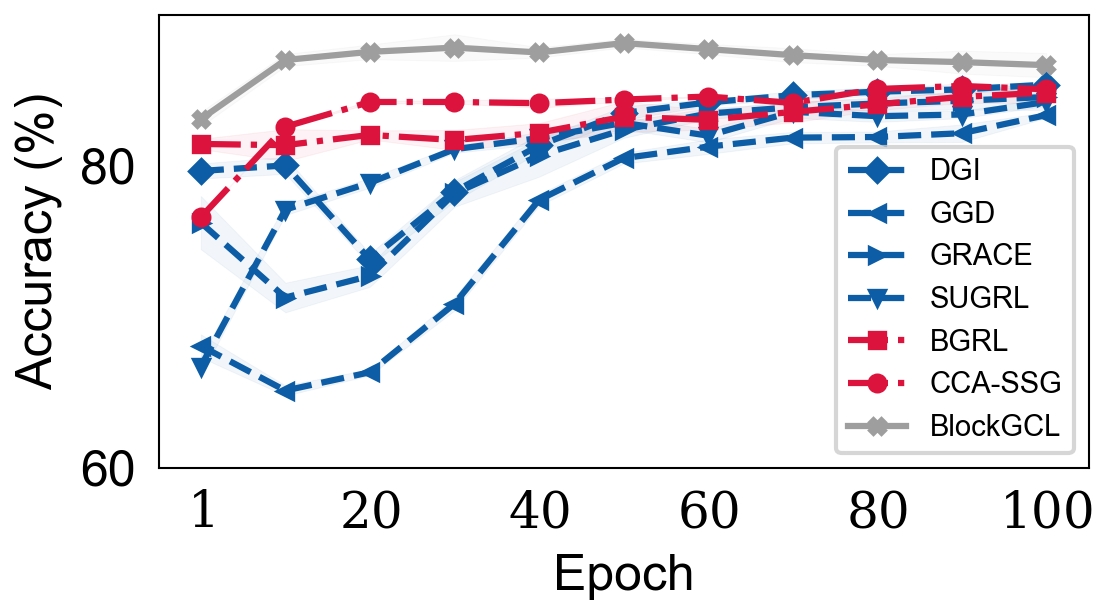}
    \caption{Empirical training curves of {\color{gray}\ours} and GCL baselines. {\color{blue2}Blue}: negative-sample-based methods; {\color{red2}red}: negative-sample-free methods.}
    \label{fig:convergence}
\end{figure}

Many fancy GCL methods require a lot of training epochs to converge, which becomes another obstacle for training large and deep GNNs and would potentially introduce the overfitting problem~\cite{bgrl,dgi}.
In this regard, we aim to investigate whether the use of blockwise training contributes to improved convergence.
We compared the convergence speed of \ours with other GCL baselines, which were categorized into negative-sample-based and negative-sample-free methods. The results on the Cora dataset are presented in Figure~\ref{fig:convergence}. Negative-sample-free methods generally converged faster, while negative-sample-based methods required more epochs to gradually improve their performance. Compared to the GCL baselines, \ours exhibits a significantly faster convergence speed, where the best performance is reached within a few epochs. We conjecture that this is likely due to the fact that our method provides guidance for each individual block without relying on a full backpropagation path from top to bottom, thus avoiding the issue of long-range starvation.
Overall, our experiments demonstrate the effectiveness of \ours in improving the convergence speed of GCL and provide insights into the benefits of the blockwise training paradigm.

\section{Conclusion}
In this work, we revisit the oversmoothing problem in the context of GCL. We first demonstrate through experiments that GCL suffers from the same oversmoothing problem as GNNs. We then investigate further how increasing network depth causes oversmoothing and discover an interesting long-range starvation phenomenon in the shallow layers of deep GCL models. Our findings suggest that without sufficient guidance from supervision, shallow representations tend to degrade and become increasingly similar. These findings motivate us to propose \ours, a simple yet effective blockwise local learning framework that enhances the robustness of deep GCL methods against oversmoothing. Extensive experimental results on five public graph benchmarks demonstrate the significant advantages of \ours against oversmoothing. Our work provides a new perspective on local learning to solve the oversmoothing problem in GCL, which would potentially inspire future research and lead to further advancements of scalable and deep GCL frameworks.

\bibliographystyle{named}
\bibliography{ijcai24}
\end{document}

\maketitle

\appendix
\section{Hyperparameters}

\begin{table}[h]
\centering
\begin{tabular}{lcc}
\toprule
\textbf{Dataset} & \textbf{Edge Dropping} & \textbf{Feature Masking} \\
\midrule
\textbf{Cora} & 0.9 & 0.4 \\
\textbf{Citeseer} & 0.6 & 0.5 \\
\textbf{Pubmed} & 0.6 & 0.7 \\
\textbf{Photo} & 0.9 & 0.1 \\
\textbf{Computers} & 0.8 & 0.2 \\
% \textbf{CS} & 0.9 & 0.2 \\
% \textbf{Physics} & 0.9 & 0\\
\bottomrule
\end{tabular}
\caption{Hyperparameters of \ours with CCA-SSG as the backbone.}\label{tab:hyperpara}

\end{table}

\section{Ablation Study}
\label{sec:abla}

\begin{figure}[t]
    \centering
    \includegraphics[width=0.48\linewidth]{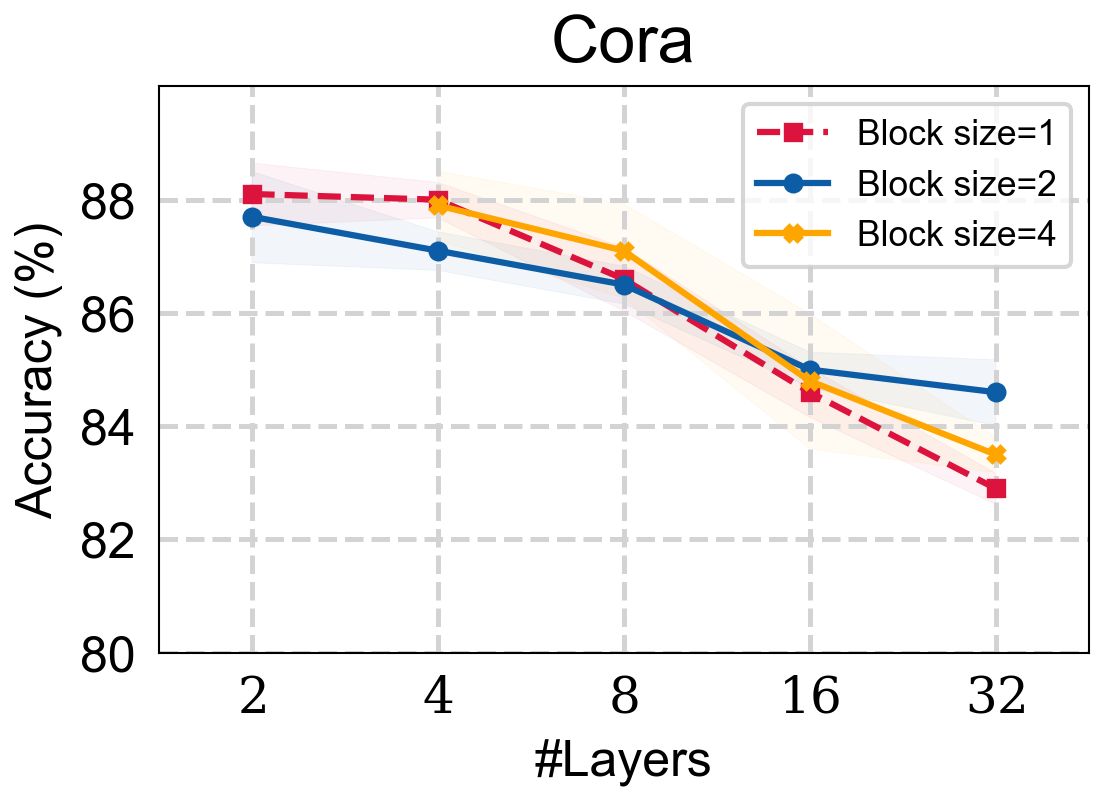}
    \includegraphics[width=0.48\linewidth]{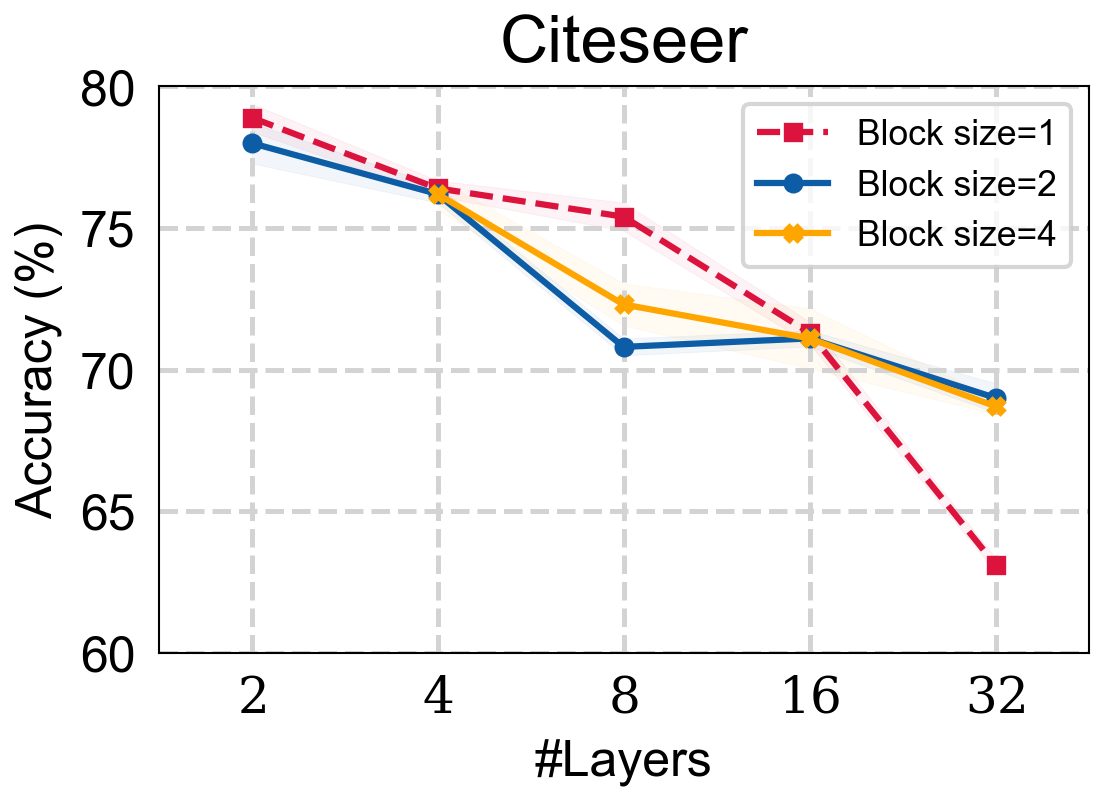}
    \includegraphics[width=0.48\linewidth]{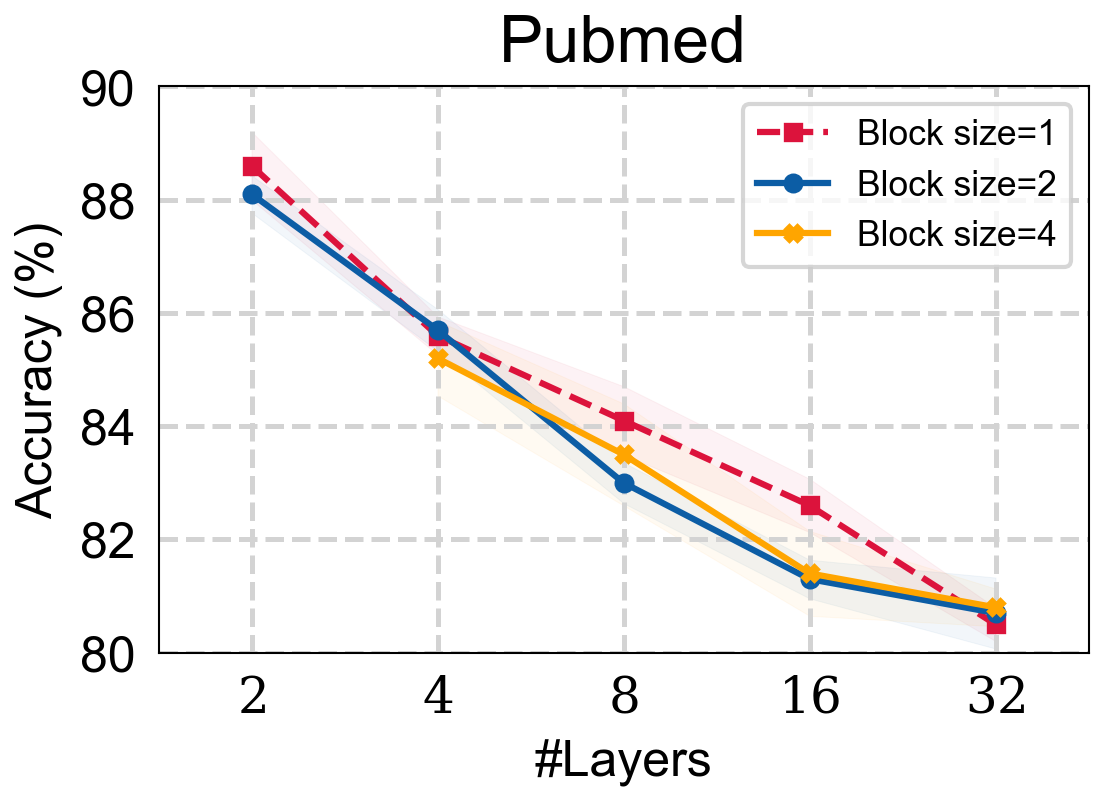}
    \caption{Performance of \ours with different blocksize and network depth.}
    \label{fig:block_size}
\end{figure}

In this section, we present an additional ablation study on block size - the depth of each block in the encoder network.
This ablation study aims to provide insights into the optimal block size for our approach.
Typically, a larger block size reduces the computational costs during loss backpropagation, and \ours degenerates to CCA-SSG when the block size is equal to the number of layers in the encoder network. To this end, we vary the block size by $\{1,2,4\}$ w.r.t. different network depth to investigate its impact. The results on three citation networks are shown in Figure~\ref{fig:block_size}. It is observed that increasing the block size in a shallow network may cause a decrease in performance, as larger blocks lose the advantage of blockwise training and may struggle to resist over-smoothing effects. In general, a block size of 1 achieves good results in most cases, while deep networks tend to favor larger blocks as their depth increases.

